\pdfoutput=1

\documentclass[11pt]{article}

\usepackage[final]{emnlp2021}

\usepackage{times}
\usepackage{latexsym}
\usepackage{hyperref}
\usepackage{bm}
\usepackage{booktabs}
\usepackage{tabularx}
\usepackage{graphicx}
\usepackage{multirow}
\usepackage{caption}
\usepackage{subcaption}

\usepackage[T1]{fontenc}

\usepackage[utf8]{inputenc}

\usepackage{microtype}

\title{Don't Discard All the Biased Instances: \\
Investigating a Core Assumption in Dataset Bias Mitigation Techniques}

\author{Hossein Amirkhani \\
  University of Qom \\
  Tehran Institute for Advanced Studies\\
  \texttt{amirkhani@qom.ac.ir} \\\And
  Mohammad Taher Pilehvar \\
  Tehran Institute for Advanced Studies, \\
  Khatam University, Iran\\
  \texttt{mp792@cam.ac.uk} \\}

\begin{document}
\maketitle
\begin{abstract}
Existing techniques for mitigating dataset bias often leverage a \textit{biased} model to identify biased instances.
The role of these biased instances is then reduced during the training of the \textit{main} model to enhance its robustness to out-of-distribution data.
A common core assumption of these techniques is that the main model handles biased instances similarly to the biased model, in that it will resort to biases whenever available.
In this paper, we show that this assumption does not hold in general. 
We carry out a critical investigation on two well-known datasets in the domain, MNLI and FEVER, along with two biased instance detection methods, partial-input and limited-capacity models. 
Our experiments show that in around a third to a half of instances, the biased model is unable to predict the main model's behavior, highlighted by the significantly different parts of the input on which they base their decisions.
Based on a manual validation, we also show that this estimate is highly in line with human interpretation.
Our findings suggest that down-weighting of instances detected by bias detection methods, which is a widely-practiced procedure, is an unnecessary waste of training data. We release our code to facilitate reproducibility and future research.\footnote{\url{https://github.com/h-amirkhani/debiasing-assumption}}
\end{abstract}

\section{Introduction}

Several studies suggest that the impressive performance of the recent natural language understanding models might not be fully indicative of the status quo in language understanding.
This is partially due to the artificial nature of the evaluation datasets which usually contain simple superficial cues, such as specific keywords or sentence structures, that spuriously correlate with the gold labels~\citep{jabri2016revisiting,jia2017adversarial,gururangan2018annotation,mccoy2019right,wiegand2019detection,schuster2019towards}.
Such shallow patterns can be easily exploited by the model, resulting in an overestimated performance on the specific dataset and usually poor performance on other differently-constructed datasets.

Many proposals have been put forward to enhance robustness to such dataset-specific biases.
Some techniques rely on the sources of bias which are known a-priori for each dataset \citep{he2019unlearn,clark2019don,mahabadi2020end}.
Others alleviate this requirement by identifying biased examples without explicitly modeling them, for instance by 
training a low-capacity model that does not go much beyond the surface patterns~\citep{sanh2020learning,utama2020towards,clark2020learning}. 
The \textit{main} model is then trained in a way not to rely much on the instances detected by the \textit{biased} model. 
This is done either implicitly using the popular \textit{product of experts} approach~\citep{clark2019don,sanh2020learning,utama2020towards,mahabadi2020end} or explicitly using methods such as \textit{debiased focal loss}~\citep{mahabadi2020end} or \textit{example reweighting}~\citep{utama2020towards}.


Irrespective of the approach for identifying biased instances, existing bias mitigation techniques share the assumption that the main model treats the biased instances similarly to the biased model, in that it will resort to superficial biased features in case they exist. We carry out a critical investigation of the validity of this assumption. 
Through a set of experiments we show that for a significant subset of instances, the biased model is unable to predict the main model's behavior.  
Specifically, we find that in around a third to a half of instances in two well-known datasets in the domain, MultiNLI \citep[MNLI]{williams2018broad} and \citep[FEVER]{thorne2018fever}, the main model handles the input differently from two popular biased instance detection methods, partial-input and limited-capacity models.
We further support this estimate through a manual validation.
This highlights the need for re-thinking the discarding of instances detected by biased models, which is a widely-adopted approach in the current dataset bias mitigation techniques.

\section{Methodology}

Following the terminology used in the literature, we refer to the model used for detecting biased instances as \textit{biased} model, the final intended model (for which we are mitigating dataset biases) as the \textit{main} model, and those instances which are correctly classified by both the main and biased models as \textit{easy} instances. 
Existing bias mitigation techniques try to weaken the role of biased instances in the learning process based on the assumption that on these instances the main model behaves similarly to the biased model. 
We investigate this assumption by comparing the parts of input (tokens) on which the two models base their decisions.
This analysis is carried out on the \textit{easy} instances where the biased model is deemed to exploit superficial features.
Specifically, we compare the role of individual input tokens in the two models to check if the main model bases its decision on the same tokens (features) as the biased model.


\paragraph{Datasets.} We experimented with two widely-used datasets in the domain: MNLI and FEVER. 
In the former, the goal is to assign each \textit{premise}-\textit{hypothesis} pair to one of three classes: \textit{entailment}, \textit{contradiction}, and \textit{neutral}; whereas in the latter, each \textit{evidence} either \textit{supports} or \textit{refutes} the corresponding \textit{claim}, or there is \textit{not enough info (NEI)} for a definite decision. 
For the MNLI data, we used \textit{validation-matched} as our validation set following~\citet{clark2019don,utama2020towards,yaghoobzadeh2021increasing}. For the FEVER data, we experimented with the version of \citet{schuster2019towards} following~\citet{mahabadi2020end,yaghoobzadeh2021increasing}. The dataset statistics are presented in Table~A1.\\


\paragraph{Main model.} For all the experiments, we fine-tuned the pre-trained BERT-base-uncased model from the Hugging Face Transformers library~\citep{wolf2019huggingface} as our main model. The hyper-parameters are chosen according to the literature~\citep{sanh2020learning}.\\

\noindent \textbf{Biased models.} We experimented with two biased instance detection (also loosely called \textit{bias-only}) methods. 
The first one is the widely-used partial-input model~\citep{gururangan2018annotation,poliak2018hypothesis,schuster2019towards}, which takes an instance as biased if an incomplete part of it is enough for correct classification.
We used the \textit{hypothesis} and \textit{claim} parts as partial inputs in the MNLI and FEVER datasets, respectively. 
The BERT-base-uncased model was trained with the same hyper-parameters as the main model on this partial-input data. 
The second method, put forward by \citet{sanh2020learning}, is based on the observation that models with limited capacity learn to exploit dataset biases. 
For this, we fine-tuned the pre-trained TinyBERT model \citep{turc2019well} with the same hyper-parameters used by \citet{sanh2020learning}.

\paragraph{Comparing biased and main models.} %
We compare the two models in terms of the input tokens on which they base their decisions.
The role of input tokens is measured using word omission, similarly to \citet{kadar2017representation}.
We consider the two models as behaving differently if their dominating input tokens differ significantly.

Consider an \textit{easy} instance $(\boldsymbol{x_i},y_i)$ which is correctly classified by both the biased $f_b$ and the main $f_m$ models. 
We denote the tokens of $\boldsymbol{x_i}$ by $\{x_{ij}\}_{j=1}^{m_i}$, where $m_i$ is the number of tokens in $\bm{x_i}$ (BERT tokenizer was used in our experiments).
The role of token $x_{ij}$ for the decision made by a classifier $f$ was computed as the change in the true class logit upon excluding that token from the input. 
More precisely, if $e_f(x_{ij})$ is the estimated effect of the token $x_{ij}$ on $f$'s decision\footnote{A better notation would be $e_f(x_{ij};\bm{x_i},y_i)$, but we omit $(\bm{x_i},y_i)$ for simplicity.}, $e_f(x_{ij})=f(\bm{x_i})_{y_i} - f(\bm{x_i}\backslash \{x_{ij}\})_{y_i}$, where $f(\bm{x})_y$ is the logit of $\bm{x}$ for the class $y$ produced by $f$. 
We represent the way that the biased model $f_b$ treats an easy instance $(\bm{x_i},y_i)$ as ${\bm{e}_{f_b}(\bm{x_i})=(e_{f_b}(x_{i1}),e_{f_b}(x_{i2}),\ldots,e_{f_b}(x_{im_i}))}$.
The same is done to obtain $\bm{e}_{f_m}(\bm{x_i})$.\footnote{Note that in the partial-input experiments, only the partial input tokens are considered in the representations.} 
Finally, cosine similarity between $\bm{e}_{f_b}(\bm{x_i})$ and $\bm{e}_{f_m}(\bm{x_i})$ is computed as an estimate for the similarity between the two models on instance $i$.
We will show in Section~\ref{sec:manual} that this estimate is highly in line with human interpretation.

\begin{figure*}
    \centering
    \includegraphics[scale=0.34]{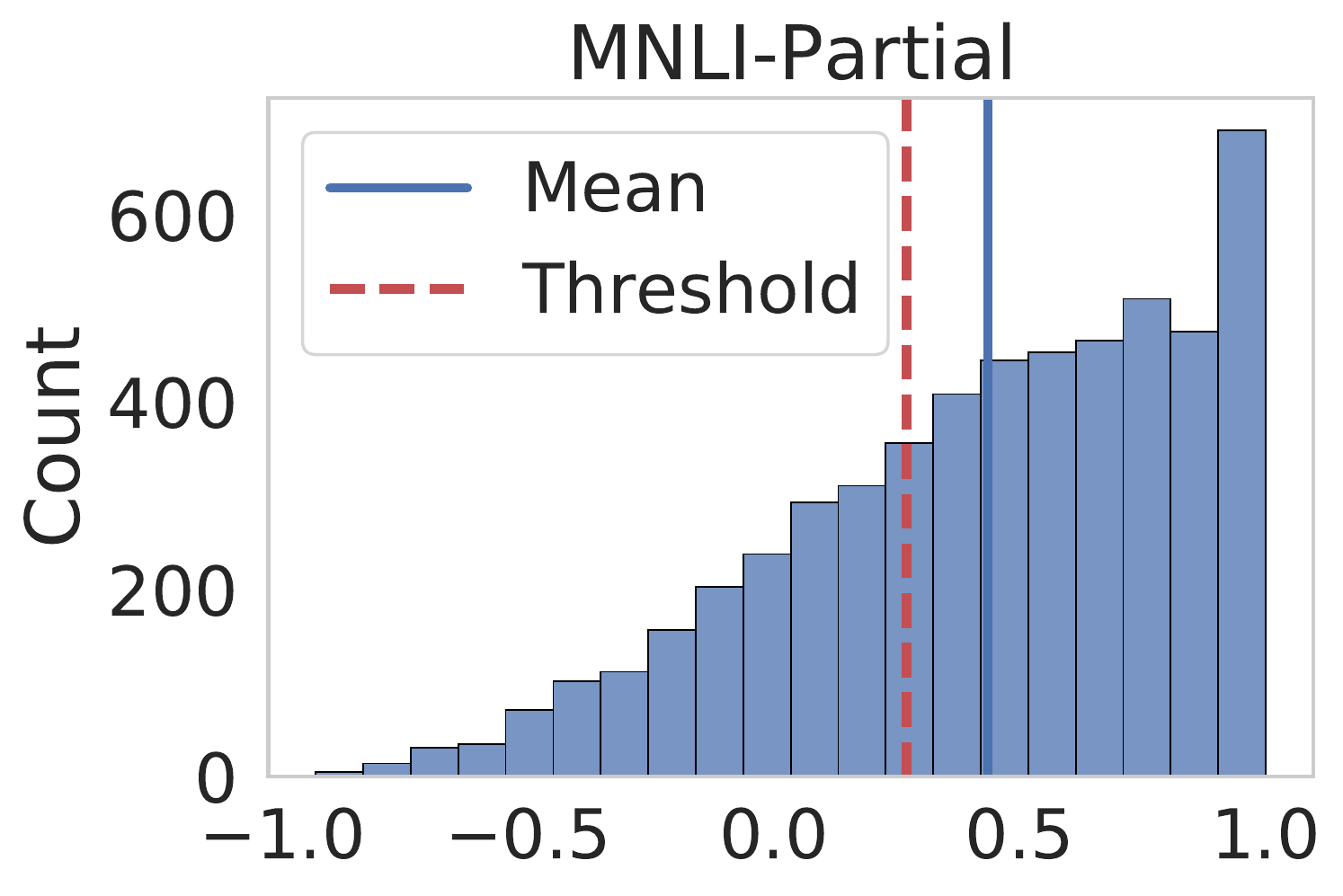}
    \includegraphics[scale=0.34]{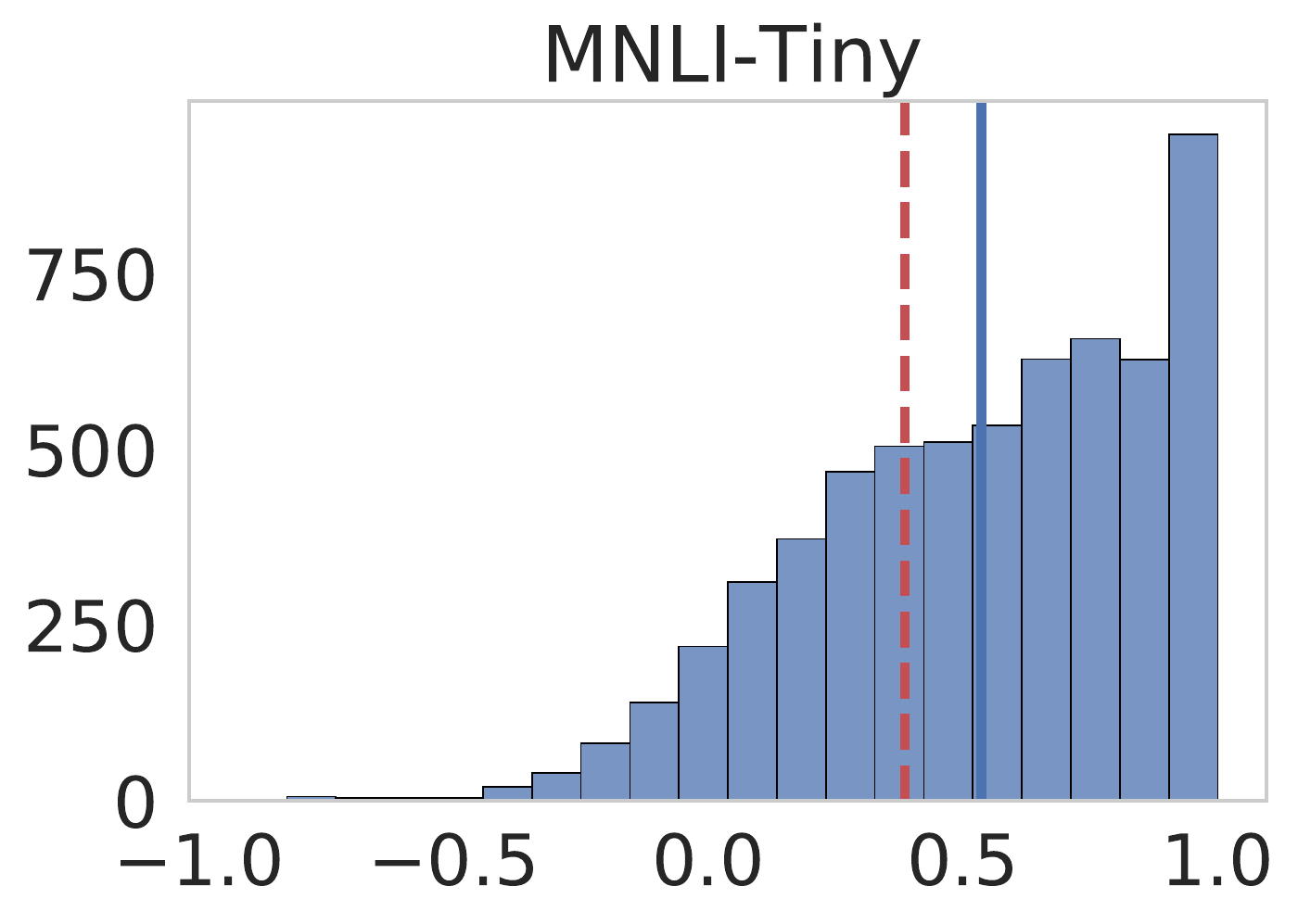}
    
    \includegraphics[scale=0.34]{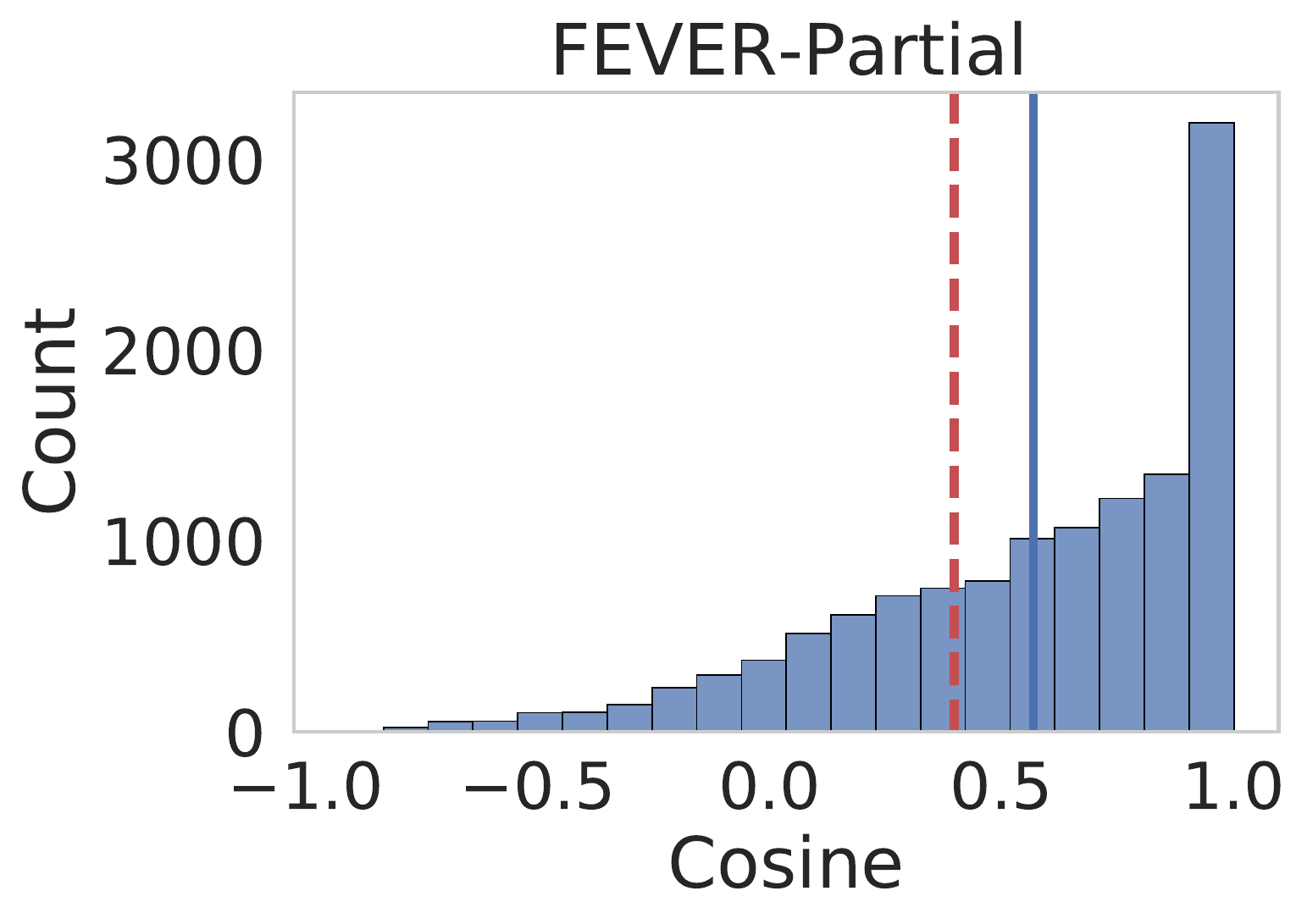}
    \includegraphics[scale=0.34]{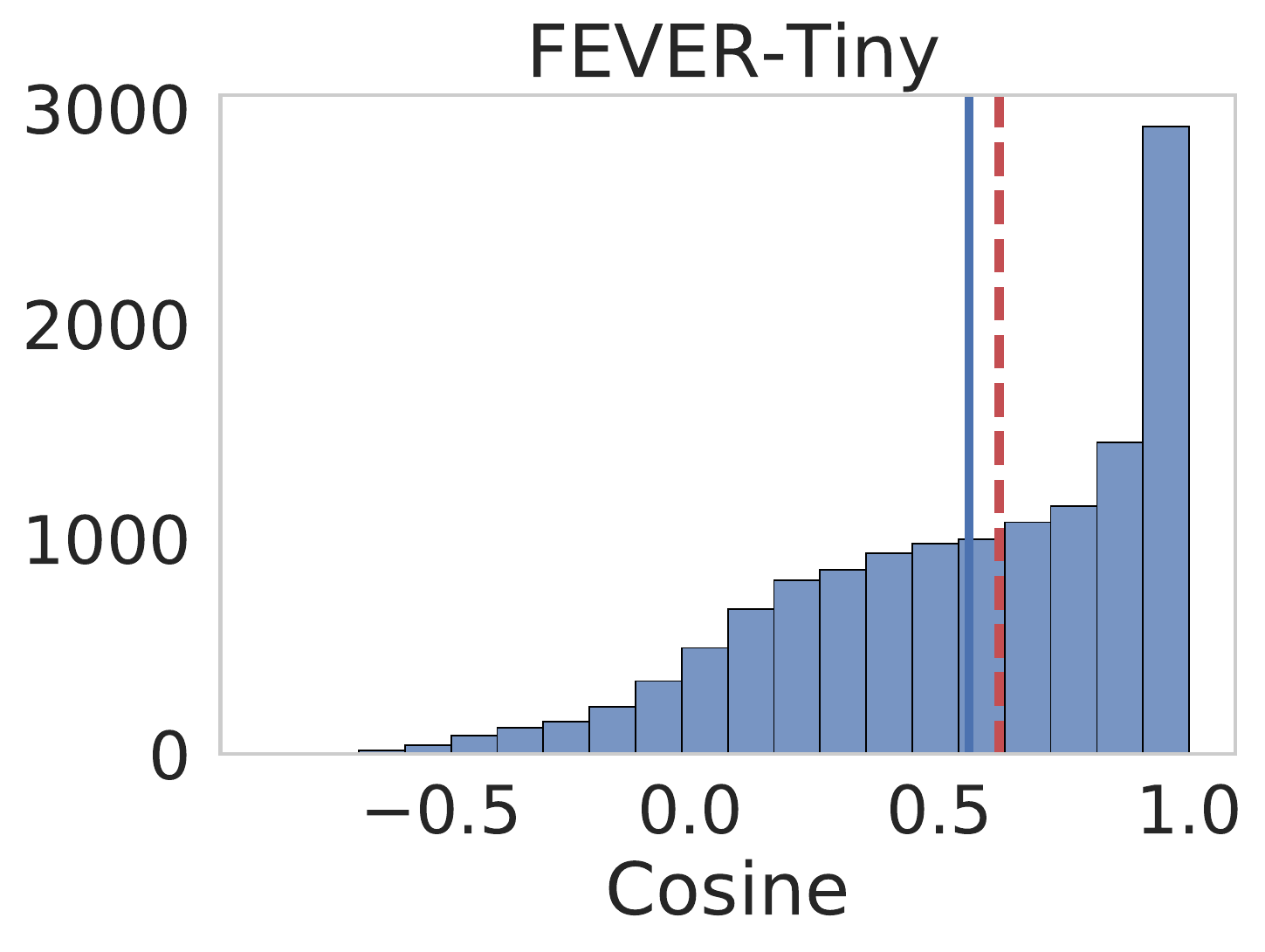}
    \caption{The distribution of inter-model (cosine) similarities. Instances below the threshold are taken as differently handled by the two models.}
    \label{fig:cosin_hist}
\end{figure*}

\section{Experiments}
 With two datasets, MNLI and FEVER,  and two biased models, partial-input and TinyBERT, there are four experimental settings in total. We refer to the two biased models as \textit{Partial} and \textit{Tiny}. 
 
\subsection{Manual validation}\label{sec:manual}
Using the methodology explained in the previous section, for each easy instance, we obtain two vectors showing the role of individual input tokens in the final decision made by the main and biased models. We hypothesize that the cosine similarity between these two vectors is correlated with the similarity of the models in treating this instance. To check this assumption, we asked two human annotators to label 250 randomly picked easy instances for each experimental setting (1000 in total). Given the omission-based vectors, the annotators' task was to judge whether the main and biased models had exploited similar evidences for their decisions or not (binary decision). 
An equal-width approach with 20 bins was used to sample instances uniformly across the cosine similarity scale (shown in Figure ~\ref{fig:cosin_hist}). 
Forty of the instances were shared among the annotators to assess their agreements.

Table~\ref{Table:manual} shows the results of this manual validation: area under the ROC curve (AUC) as well as Inter-annotator Agreement (IAA). AUC is the probability that a randomly chosen \textit{positive} instance obtains a higher cosine similarity than a randomly chosen \textit{negative} one, where the positive class denotes those instances  which are treated similarly by the two models according to the annotators. 
High AUC values for all the four settings show that the cosine similarity between omission-based vectors of main and biased models is highly in line with the human perception of the similarity between behavior of the two models. 
The high inter-annotator agreement across all the four settings confirms the reliability of this manual validation. 

    \begin{table}[t!]
	\setlength{\tabcolsep}{12.0pt}
	\scalebox{0.9}{
		\begin{tabular}{llcc}
		\toprule
		  Dataset & Biased model & IAA & AUC \\
		  \midrule
		  \multirow{2}{*}{MNLI} & Partial & 0.900 & 0.974\\ 
		  & Tiny & 0.875 & 0.965\\
		  \midrule
		  \multirow{2}{*}{FEVER} & Partial & 0.900 & 0.969\\
		  & Tiny & 0.875  & 0.925\\
		  \bottomrule
		\end{tabular}
		}
		\caption{Manual validation of the hypothesis that cosine similarity between omission-based vectors obtained from the main and biased models can represent how similarly they treat instances. We report Inter-annotator Agreement (IAA, in terms of accuracy) along with Area Under Curve (AUC).}
		\label{Table:manual}
	\end{table}

\begin{table}[t!]
	\setlength{\tabcolsep}{5pt}
	\scalebox{0.9}{
		\begin{tabular}{lllll}
		\toprule
		  Dataset & Bias & Easy & F1 & Different \\
		  \midrule
		  \multirow{2}{*}{MNLI} & Partial & 5,381 (54.8) & 93.4 & 1,723 (32.0)\\ 
		  & Tiny & 6,090 (62.0) & 94.0 & 1,973 (32.4)\\
		  \midrule
		  \multirow{2}{*}{FEVER} & Partial & 12,627 (63.1) & 94.5 & 3,794 (30.0)\\
		  & Tiny & 13,328 (66.6) & 90.9 & 6,599 (49.5)\\
		  \bottomrule
		\end{tabular}
		}
		\caption{The total number of \textit{easy} instances (\% in parentheses) and the subset identified as being treated differently by the two models (\textit{different}). We also report the classifier's F1 on the gold data.}
		\label{Table:quantitative}
	\end{table}

\begin{figure*}[t!]
    \centering
        \includegraphics[trim=0 0 0 10,clip,width=1.0\linewidth]{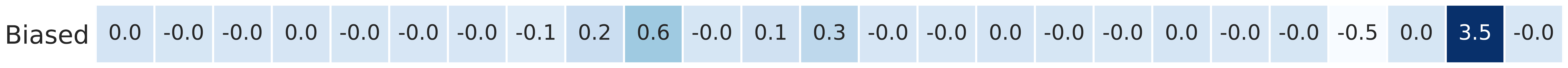}
        \includegraphics[trim=0 0 0 0,clip,width=1.0\linewidth]{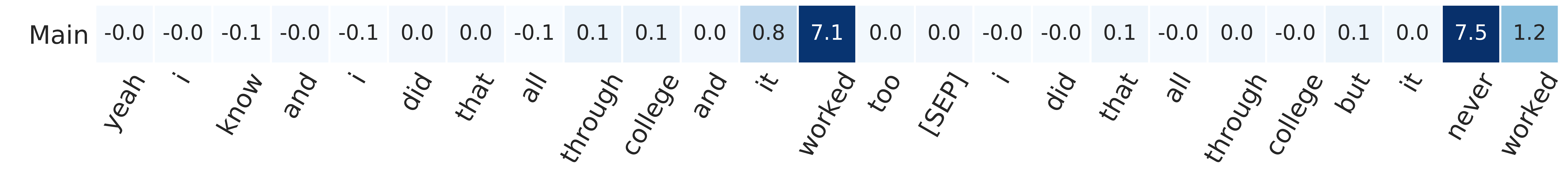}
    
        \includegraphics[trim=0 0 0 0,clip,width=1.0\linewidth]{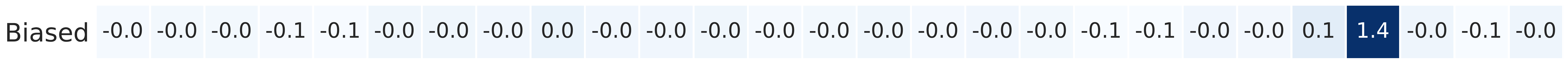}
        \includegraphics[trim=0 0 0 0,clip,width=1.0\linewidth]{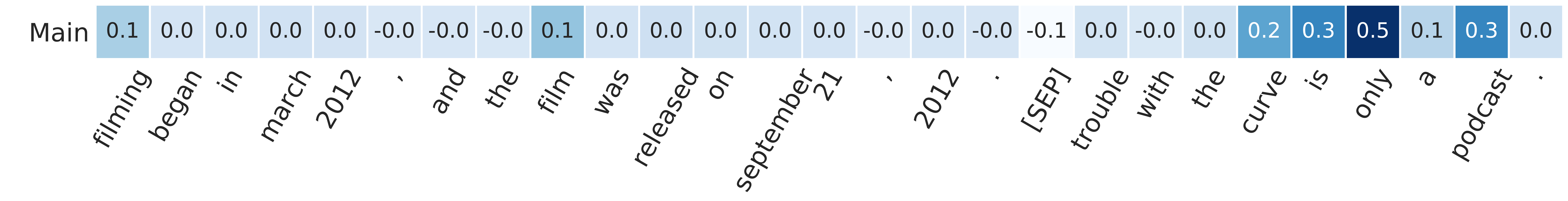}
    
    \caption{Examples of \textit{different} instances from MNLI (top) and FEVER (bottom) datasets.}
    \label{fig:example_work_film}
\end{figure*}


\subsection{Quantitative analysis}

To have an estimate on the number of easy instances that were handled differently by the two models, we used a simple threshold-based binary classifier tuned based on F1-score on the negative instances in the manually labeled data.
The computed thresholds are shown in Figure \ref{fig:cosin_hist} as red (dashed) lines. 
Instances with an inter-model similarity below this threshold are regarded as being differently treated by the two models.


The results of the quantitative analysis are presented in Table~\ref{Table:quantitative}. 
The \textit{easy} column shows the number (and the percentage in parentheses) of easy instances, i.e., those that are correctly classified by both the main and biased models. 
The subset of easy instances which are identified by the threshold-based classifier as being differently treated by the two models is shown under column \textit{different}.
We also report the F1-score for the corresponding classifier.
The high F1-scores indicate that the threshold-based classifier is accurate in predicting the gold labels assigned by the annotators. 

The results show that the main model handles a considerable subset of easy instances differently from the biased model. This undermines the soundness of a core assumption made by many bias mitigation techniques.
Specifically, for three of the four settings, this comprises around a third of easy instances, the subset which is often unnecessarily discarded by the bias mitigation techniques.
The FEVER-Tiny setting is an exception for which the estimate is close to 50\% of the easy instances. This can explain the particular brittleness of TinyBERT-based bias detection approach on the FEVER dataset which explains the lower performance of this approach reported by \citet{sanh2020learning} compared to partial-input.

\begin{figure*}[t!]
    \centering
    \includegraphics[trim=0 0 0 20,clip,width=0.7\linewidth]{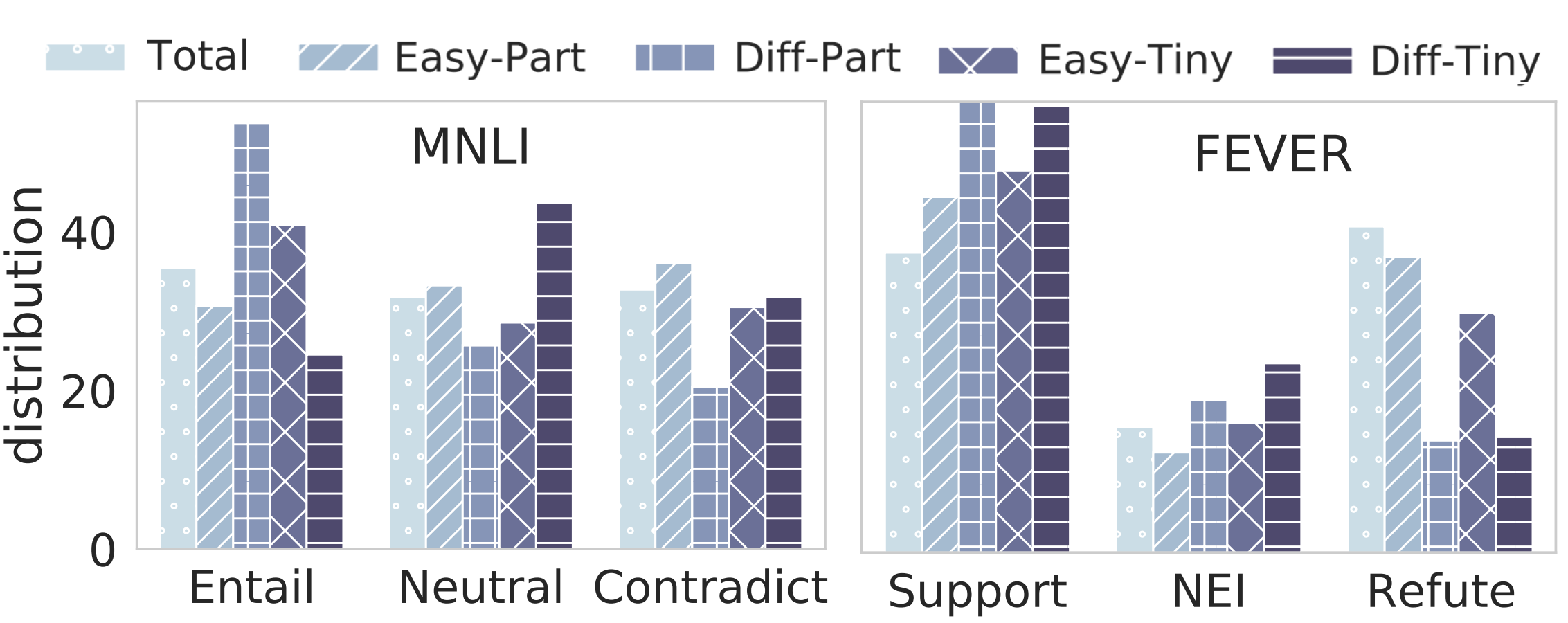}
    \caption{Label distribution in the validation data (total), easy instances (Easy-), and the subset identified as being handled differently by the two models (Diff-) and for the two bias detection models (-Part and -Tiny).}
    \label{fig:settings_dist}
\end{figure*}

\subsection{Discussion} 

We showed that the main model does not follow the biased models for a significant number of instances. Figure \ref{fig:example_work_film} shows two such instances.
On top, the biased model seems to have based its decision mostly on the negative word \textit{never} in the hypothesis to make a \textit{contradiction} prediction.
Despite having access to this keyword, the main model utilizes a wider range of relevant words such as \textit{worked} in the premise and the negated phrase \textit{never worked} in the hypothesis.
The same is true for the second example where the biased model uses the negative word \textit{only} whereas the main model takes additional evidence such as the contradiction between \textit{film} in the premise and \textit{podcast} in the hypothesis. More examples are shown in the appendix.

Figure~\ref{fig:settings_dist} shows the distribution of \textit{easy} and \textit{different} instances over the three dataset labels. 
In FEVER, a significant proportion of \textit{different} samples belong to the \textit{support} class, while the \textit{refute} class has the smallest share.\footnote{In this section, we use the MNLI and FEVER terminology interchangeably.}  
The same holds for the \textit{MNLI-Partial} setting. This shows that in most cases, it is much more likely for the main model to find evidences that are different from those exploited by the biased model when the \textit{premise} entails \textit{hypothesis}.
On the other hand, when the premise contradicts the hypothesis, the main model highly resorts to the same cues detected by the biased model. 

This observation suggests that class labels are helpful information which can be exploited by bias mitigation techniques. 
In other words, instance weights can be determined not only based on the biased model's loss during training, but also by incorporating the corresponding classes. 
For example, it is better to reduce the down-weighting for those instances in the FEVER dataset that belong to the \textit{support} class.

\section{Related Work}
The presence of dataset-specific superficial cues have been shown in different NLP tasks including visual question answering~\citep{jabri2016revisiting,manjunatha2019explicit}, machine reading comprehension~\citep{jia2017adversarial}, natural language inference~\citep{gururangan2018annotation,poliak2018hypothesis}, abusive language detection~\citep{wiegand2019detection}, and fact verification~\citep{schuster2019towards}.
Models that rely on these dataset biases usually have low performance in out-of-distribution settings. To measure the reliance on non-generalizable patterns, various challenging sets have been proposed ~\cite{jia2017adversarial,agrawal2018don,mccoy2019right,zhang2019paws,sakaguchi2020winogrande}. There have also been some attempts to systematically reduce dataset biases during construction~\cite{zellers2018swag,le2020adversarial}. Despite some bias reduction reported, the datasets may still contain hidden biased patterns~\cite{sharma2018tackling}; therefore, it is crucial to empower the learning algorithms to be robust against biases.

A popular approach to mitigate dataset biases is to encourage the main model to pay less attention to the instances that are correctly classified by a biased model. To train a biased model, some methods use a-priori known sources of biases. For instance, good performance given insufficient semantics of the input is attributed to bias exploitation~\citep{he2019unlearn,clark2019don,cadene2019rubi,mahabadi2020end}. Others try to identify biased instances without explicitly modeling them, such as by training a limited capacity model~\cite{sanh2020learning,clark2020learning} or exposing the model to only a small number of training instances~\cite{utama2020towards}. To decrease the reliance of the model on the (likely) biased instances, some techniques implicitly reduce the updates of main model's parameters for biased instances ~\cite{clark2019don,cadene2019rubi,sanh2020learning,utama2020towards,mahabadi2020end}. 
Others explicitly downweight the biased instances, for instance using debiased focal loss~\cite{mahabadi2020end} or example reweighting~\cite{utama2020towards}. 

\section{Conclusions}

Through a set of experiments, we showed that a common core assumption of dataset bias mitigation methods does not hold for a significant portion of two widely-used benchmarks.
Specifically, we observed that two widely-used bias detection approaches, partial-input and low-capacity model, are unable to accurately predict model's handling of biased instances.
We carried out extensive analysis and manual validation to attest the reliability of this observation.
We infer that what identifies a biased instance is not the instance itself, but the way the model treats it. 
In other words, an instance that has evident biased patterns is not necessarily useless as long as the main model does not base its decision on these biases.

It is worth noting that the dissimilarity in the handling of input between the biased and main models does not imply that the latter necessarily adopts an unbiased strategy. 
In other words, it is possible for the main model to treat the instance differently from the biased model but still exploits a (different) bias. 
As immediate future work, we plan to build on the findings of this analysis for enhancing dataset bias mitigation techniques. 


\bibliographystyle{acl_natbib}
\bibliography{emnlp2021}

\appendix
\setcounter{table}{0}
\renewcommand{\thetable}{\thesection\arabic{table}}
\setcounter{figure}{0} 
\renewcommand\thefigure{\thesection\arabic{figure}}    

\section{Appendix}
\label{sec:appendix}

\subsection{Experimental setup}
Table~\ref{Table:datasets} shows the statistics of datasets used in experiments. 
For the FEVER data, we experimented with the version of \citet{schuster2019towards}\footnote{\url{https://www.dropbox.com/s/v1a0depfg7jp90f/fever.train.jsonl} \newline \indent \indent\url{https://www.dropbox.com/s/bdwf46sa2gcuf6j/fever.dev.jsonl}} augmented with the \textit{NEI} samples used in~\citet{Schuster2021}\footnote{\url{https://github.com/TalSchuster/talschuster.github.io/raw/master/static/vitaminc_baselines/fever.zip}}.

Following~\citet{sanh2020learning}, the models are trained for three epochs with a learning rate of 2e-5 and a batch size of 8. The weight decay rate used for MNLI and FEVER are 0.01 and 0.1, respectively. For FEVER, the learning rate is linearly increased for 1000 warming steps and linearly decreased to 0 afterward. Other hyper-parameters are left as default.

For manual validation, the authors labeled 250 randomly picked easy instances for each experimental setting. They were shown the word omission-based role of different tokens in the main and biased models (as in Figure~\ref{fig:more_examples}) to decide whether their dominating input tokens differed significantly. As the control check, the Inter-annotator Agreement (IAA) was monitored.

    \begin{table}[ht!]
    \centering
	\setlength{\tabcolsep}{4.0pt}
	\scalebox{0.85}{
		\begin{tabular}{l l r ccc}
		\toprule
		  & & \# & Entail & Neutral & Contradict \\
		  \midrule
		  \multirow{2}{*}{MNLI} & train & 392,702 & 33.3\% & 33.3\% & 33.3\% \\ & valid. & 9,815 & 35.5\% & 31.8\% & 32.7\%\\
		  \midrule
		  \multirow{2}{*}{FEVER} & train & 242,911 & 41.4\% & 41.4\% & 17.2\% \\
		  & valid. & 19,997 & 39.9\% & 16.7\% & 43.4\% \\
		  \bottomrule
		\end{tabular}
		}
		\caption{The datasets used in the experiments.}
		\label{Table:datasets}
	\end{table}

\subsection{Additional examples}
\begin{figure*}[ht!]
    \centering
        \begin{subfigure}[b]{\textwidth}
            \includegraphics[width=\linewidth]{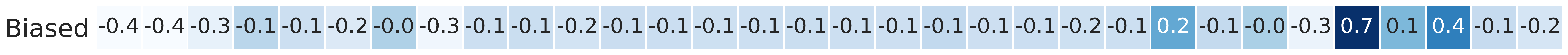}
            \includegraphics[width=\linewidth]{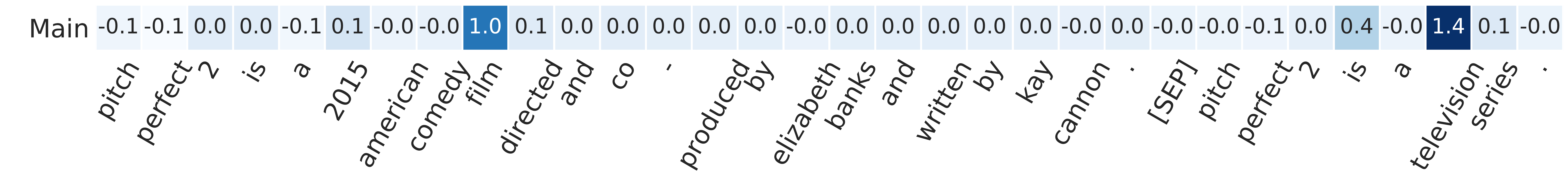}
            \caption{An instance from the FEVER dataset. Note that the contradiction between \textit{film} and \textit{television series} is correctly exploited by the main model which is missed by the Tiny biased model.}
        \end{subfigure}

        \begin{subfigure}[b]{\textwidth}
            \includegraphics[trim=0 0 0 -200, width=\linewidth]{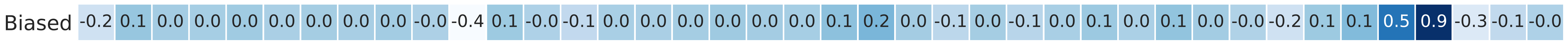}
            \includegraphics[width=\linewidth]{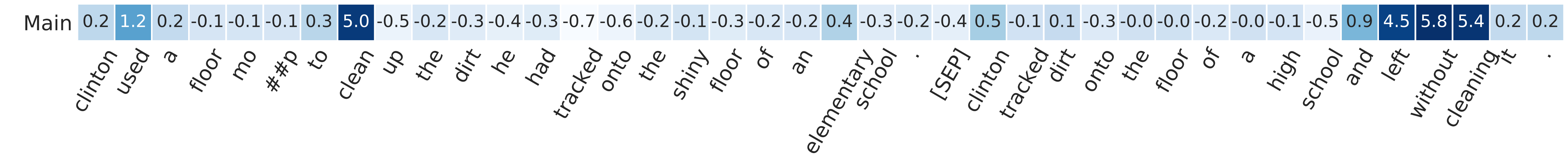}
            \caption{An instance from the MNLI dataset. The main model has recognized the contradiction between \textit{clean} in the premise and \textit{left without cleaning} in the hypothesis which is missed by the Tiny biased model.}
        \end{subfigure}
        
        \begin{subfigure}[b]{\textwidth}
            \includegraphics[trim=0 0 0 -200,width=\linewidth]{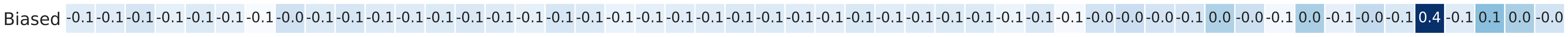}
           \includegraphics[width=\linewidth]{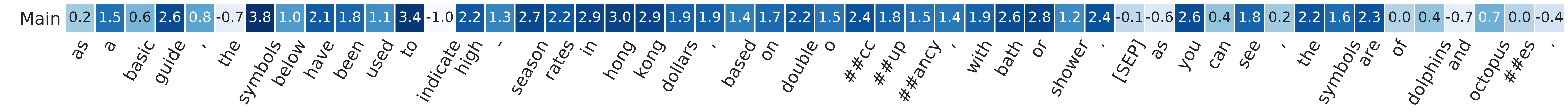}
            \caption{A neutral instance from the MNLI dataset. The biased model based its decision on a single word, while the main model exploits a wide range of words.}
        \end{subfigure}
        
        \begin{subfigure}[b]{\textwidth}
        \centering
            \includegraphics[trim=0 0 0 -100, width=0.7\linewidth]{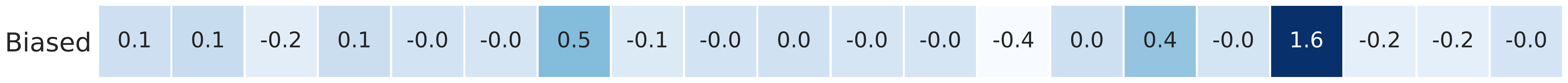}
            \includegraphics[width=0.7\linewidth]{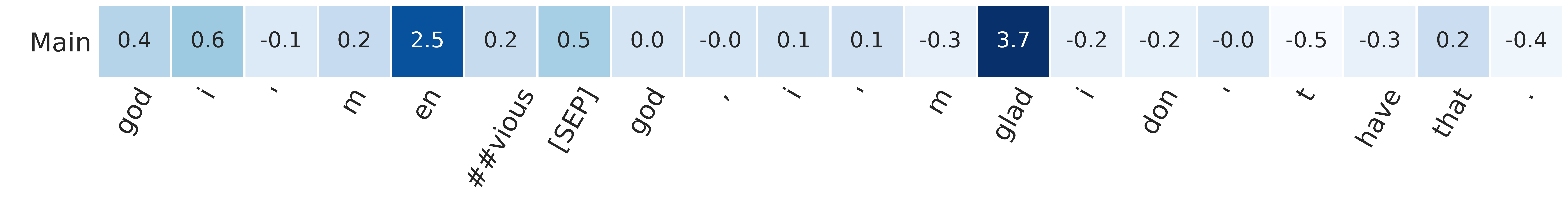}
            \caption{An instance from the MNLI dataset. The main model exploits the contradiction between \textit{envious} and \textit{glad}, while the biased model just relies on the negative verb \textit{don't} to classify it as contradiction.}
        \end{subfigure}
    \caption{More examples of instances which are treated differently by the biased and main models.}
    \label{fig:more_examples}
\end{figure*}

Figure ~\ref{fig:more_examples} shows some more examples of the instances which are treated differently by the biased and main models. 

\end{document}